\documentclass{article}

\PassOptionsToPackage{numbers, sort&compress, authoryear}{natbib}
\usepackage[preprint]{neurips_2025}

\usepackage[utf8]{inputenc} 
\usepackage[T1]{fontenc}    
\usepackage{xcolor}         
\definecolor{cvprblue}{rgb}{0.21,0.49,0.74}
\usepackage[pagebackref,breaklinks,colorlinks,allcolors=cvprblue]{hyperref}
\usepackage{graphicx}
\usepackage{url}            
\usepackage{booktabs}       
\usepackage{amsfonts}       
\usepackage{nicefrac}       
\usepackage{microtype}      
\usepackage{fontawesome5}
\usepackage{amsmath}
\usepackage{colortbl,xcolor}
\usepackage{amssymb}
\usepackage{booktabs}
\usepackage{multirow}
\usepackage{colortbl}
\usepackage{subcaption}
\usepackage{wrapfig}
\usepackage{multicol}

\newlength\savewidth\newcommand\shline{\noalign{\global\savewidth\arrayrulewidth
  \global\arrayrulewidth 1pt}\hline\noalign{\global\arrayrulewidth\savewidth}}

\newcommand{\tablestyle}[2]{\setlength{\tabcolsep}{#1}\renewcommand{\arraystretch}{#2}\centering\footnotesize}

\renewcommand{\paragraph}[1]{\vspace{0mm}\noindent\textbf{#1}}
\definecolor{gray}{gray}{0.95}
\definecolor{lightyellow}{rgb}{1.0, 0.98, 0.8}
\definecolor{lightblue}{rgb}{0.85, 0.9, 1.0}
\definecolor{lightgreen}{rgb}{0.85, 1.0, 0.9}
\definecolor{hardlabel}{HTML}{e5e5ff}
\definecolor{softlabel}{HTML}{fee5e5}

\title{DD-Ranking: Rethinking the Evaluation of Dataset Distillation}

\author{%
  DD-Ranking Team\thanks{See all members in Appendix~\ref{dd_ranking_team}. Zhiwei Deng (Google DeepMind) served as an external advisor only.} \\
  \faGithub \ \url{https://github.com/NUS-HPC-AI-Lab/DD-Ranking.git} \\
  \faBook \ \url{https://nus-hpc-ai-lab.github.io/DD-Ranking/} \\
  \faMedal \ \url{https://huggingface.co/spaces/logits/DD-Ranking} 
}

\begin{document}

\maketitle

\begin{abstract}
In recent years, dataset distillation has provided a reliable solution for data compression, where models trained on the resulting smaller synthetic datasets achieve performance comparable to those trained on the original datasets. To further improve the performance of synthetic datasets, various training pipelines and optimization objectives have been proposed, greatly advancing the field of dataset distillation. Recent decoupled dataset distillation methods introduce soft labels and stronger data augmentation during the post-evaluation phase and scale dataset distillation up to larger datasets (\textit{e.g.}, ImageNet-1K). However, this raises a question: \textit{Is accuracy still a reliable metric to fairly evaluate dataset distillation methods?} 
Our empirical findings suggest that the performance improvements of these methods often stem from additional techniques rather than the inherent quality of the images themselves, with even randomly sampled images achieving superior results. Such misaligned evaluation settings severely hinder the development of DD. Therefore, we propose DD-Ranking, a unified evaluation framework, along with new general evaluation metrics to uncover the true performance improvements achieved by different methods. By refocusing on the actual information enhancement of distilled datasets, DD-Ranking provides a more comprehensive and fair evaluation standard for future research advancements. 
\end{abstract}
\section{Introduction}
With the rapid advancement of deep learning, training increasingly complex and more complex models on large scale datasets has become a standard paradigm, achieving remarkable performance in various fields, such as computer vision~\cite{he2016deep,dosovitskiy2020image} and natural language processing~\cite{devlin2018bert,brown2020language}. However, this process often incurs substantial computational and storage demands, significantly hindering deployment across diverse scenarios. Dataset distillation (DD)~\cite{wang2018dataset}, as a recent promising solution for dataset compression, offers novel insights to address these challenges. In recent years, diverse training pipelines~\cite{kim2022dataset, NEURIPS2022_de3d2bb6, zhao2022synthesizing, feng2024embarrassingly, yin2024squeeze} and optimization objectives~\cite{zhao2021dataset,zhao2021datasetdm,cazenavette2022dataset} have been proposed, driving rapid advancement in the field of dataset distillation.

To further enhance the testing accuracy of models trained on synthetic datasets during the post-evaluation phase, recent studies have incorporated general performance boosting techniques (e.g., soft labels) into the evaluation process. Some methods jointly optimize the generated images and their corresponding unique soft labels~\cite{guo2023towards,loo2022efficient}, while decoupled dataset distillation methods~\cite{yin2024squeeze,shao2024generalized,su2024d4m,sun2024diversity,qin2024label} utilize epoch-wise soft labels provided by pre-trained teacher models during post-evaluation phase. Although these works successfully demonstrate that soft labels significantly improve testing accuracy of the validation models, their soft label implementation strategies differ substantially, and performance comparisons with prior methods often fail to account for gains attributable to soft labels. 

Furthermore, subsequent studies frequently employ more intensive data augmentation, superior optimizers, and refined training hyper-parameters~\cite{shao2024elucidating,cui2025dataset} during evaluation to maximize model performance, with even randomly sampled images achieving superior results under better post-evaluation settings. This practice conflates genuine improvements in dataset quality with performance variations caused by inconsistent evaluation settings, severely impeding progress in dataset distillation and directing subsequent improvements toward suboptimal directions.
Based on the aforementioned discussion, we must emphasize that in the growing field of dataset distillation, relying solely on the testing accuracy of validation model as the exclusive criterion for assessing the quality of synthetic datasets exhibits significant unreliability and unfairness when applied across varying settings. 

To address these issues, we propose DD-Ranking, a unified evaluation framework, and introduce a new fair and generalizable metric to realign with the original objectives of dataset distillation. Specifically, we first test evaluation models using randomly sampled images under the evaluation settings of various distillation methods to establish baseline performance for different settings. The performance of generated images is then calibrated by calculating the difference from this baseline. On the other hand, we compute the difference between the performance of synthetic datasets under the hard label settings and the maximum achievable performance using the full original dataset. By jointly applying these two adaptive metrics to evaluate existing distillation methods, we derive a new performance indicator that reveals the true differences in distillation capabilities among methods. Building upon this, we also propose a novel metric for evaluating data augmentations. We further examine the robustness of the introduced metrics across diverse application scenarios. 


DD-Ranking addresses the inconsistencies present in existing dataset distillation evaluation protocols and unifies various methods under a fair and standardized evaluation framework, thereby establishing a solid baseline and offering valuable insights for future research. The contributions of our benchmark are threefold. First, we standardize evaluation metrics for dataset distillation, resolving the persistent issue of unfair comparisons in test accuracy across different methods. Second, experimental observations from DD-Ranking demonstrate that previous performance improvements commonly originate from the enhanced model training techniques instead of the distilled dataset.
Thus, DD-Ranking encourages the community to direct future efforts toward enhancing the informativeness of synthetic data. Third, building upon the era of dataset distillation, we introduce a general and robust metric that serves as a novel evaluation criterion, with broader applicability across data-centric AI tasks.
\begin{table}[t]
\centering
\tablestyle{4pt}{1.2}
\resizebox{\textwidth}{!}{%
\begin{tabular}{c|ccccccccccccc}
Config & DC   & DSA  & DM   & MTT  & DataDAM & DATM   & SRe2L    & RDED     & CDA      & DWA     & D4M      & EDC      & G-VBSM   \\ \shline
Epoch                   & \cellcolor[HTML]{aac9f9}{1K} & \cellcolor[HTML]{aac9f9}{1K} & \cellcolor[HTML]{aac9f9}{1K} & \cellcolor[HTML]{aac9f9}{1K} & \cellcolor[HTML]{aac9f9}{1K} & \cellcolor[HTML]{aac9f9}{1K}  & \cellcolor[HTML]{c6f4c6}{300}      & \cellcolor[HTML]{c6f4c6}{300}   & \cellcolor[HTML]{c6f4c6}{300}      & \cellcolor[HTML]{c6f4c6}{300}   & \cellcolor[HTML]{c6f4c6}{300}    & \cellcolor[HTML]{c6f4c6}{300}   & \cellcolor[HTML]{c6f4c6}{300}      \\
Batch Size   & \cellcolor[HTML]{aac9f9}{256}  & \cellcolor[HTML]{aac9f9}{256}  & \cellcolor[HTML]{aac9f9}{256}  & \cellcolor[HTML]{aac9f9}{256}  & \cellcolor[HTML]{aac9f9}{256}     & \cellcolor[HTML]{aac9f9}{256}    & \cellcolor[HTML]{c6f4c6}{1024}     & \cellcolor[HTML]{fbecc9}{100}      & \cellcolor[HTML]{fccece}{128}       & \cellcolor[HTML]{fccece}{128}     & \cellcolor[HTML]{c6f4c6}{1024}     & \cellcolor[HTML]{fbecc9}{100}      & \cellcolor[HTML]{c6f4c6}{1024}      \\
Optimizer  & \cellcolor[HTML]{aac9f9}{SGD}  & \cellcolor[HTML]{aac9f9}{SGD}  & \cellcolor[HTML]{aac9f9}{SGD}  & \cellcolor[HTML]{aac9f9}{SGD}  & \cellcolor[HTML]{aac9f9}{SGD}     & \cellcolor[HTML]{aac9f9}{SGD}    & \cellcolor[HTML]{c6f4c6}{AdamW}    & \cellcolor[HTML]{c6f4c6}{AdamW}    & \cellcolor[HTML]{c6f4c6}{AdamW}    & \cellcolor[HTML]{c6f4c6}{AdamW}   & \cellcolor[HTML]{c6f4c6}{AdamW}    & \cellcolor[HTML]{c6f4c6}{AdamW}    & \cellcolor[HTML]{c6f4c6}{AdamW}   \\
LR Scheduler & \cellcolor[HTML]{aac9f9}{step} & \cellcolor[HTML]{aac9f9}{step} & \cellcolor[HTML]{aac9f9}{step} & \cellcolor[HTML]{aac9f9}{step} & \cellcolor[HTML]{aac9f9}{step}  & \cellcolor[HTML]{aac9f9}{step} & \cellcolor[HTML]{c6f4c6}{cosine}   & \cellcolor[HTML]{c6f4c6}{cosine}  & \cellcolor[HTML]{c6f4c6}{cosine} & \cellcolor[HTML]{c6f4c6}{cosine} & \cellcolor[HTML]{c6f4c6}{cosine}  & \cellcolor[HTML]{c6f4c6}{cosine}  & \cellcolor[HTML]{c6f4c6}{cosine}  \\
Label Type   & \cellcolor[HTML]{aac9f9}{hard} & \cellcolor[HTML]{aac9f9}{hard} & \cellcolor[HTML]{aac9f9}{hard} & \cellcolor[HTML]{aac9f9}{hard} & \cellcolor[HTML]{aac9f9}{hard}  & \cellcolor[HTML]{c6f4c6}{soft}  & \cellcolor[HTML]{c6f4c6}{soft}  & \cellcolor[HTML]{c6f4c6}{soft}  & \cellcolor[HTML]{c6f4c6}{soft}  & \cellcolor[HTML]{c6f4c6}{soft}  & \cellcolor[HTML]{c6f4c6}{soft} & \cellcolor[HTML]{c6f4c6}{soft} & \cellcolor[HTML]{c6f4c6}{soft}  \\
Soft Label              & \cellcolor[HTML]{fccece}{-}    & \cellcolor[HTML]{fccece}{-}    & \cellcolor[HTML]{fccece}{-}   & \cellcolor[HTML]{fccece}{-}    & \cellcolor[HTML]{fccece}{-}  & \cellcolor[HTML]{aac9f9}{single} & \cellcolor[HTML]{c6f4c6}{multiple} & \cellcolor[HTML]{c6f4c6}{multiple} & \cellcolor[HTML]{c6f4c6}{multiple} & \cellcolor[HTML]{c6f4c6}{multiple} & \cellcolor[HTML]{c6f4c6}{multiple} & \cellcolor[HTML]{c6f4c6}{multiple} & \cellcolor[HTML]{c6f4c6}{multiple} \\
Loss Function           & \cellcolor[HTML]{aac9f9}{CE} & \cellcolor[HTML]{aac9f9}{CE} & \cellcolor[HTML]{aac9f9}{CE} & \cellcolor[HTML]{aac9f9}{CE} & \cellcolor[HTML]{aac9f9}{CE} & \cellcolor[HTML]{fbecc9}{SCE}   & \cellcolor[HTML]{dbdcf9}{KL}  & \cellcolor[HTML]{dbdcf9}{KL}   & \cellcolor[HTML]{dbdcf9}{KL}  & \cellcolor[HTML]{dbdcf9}{KL}   & \cellcolor[HTML]{dbdcf9}{KL}  & \cellcolor[HTML]{c6f4c6}{MSE}  & \cellcolor[HTML]{c6f4c6}{MSE}      \\
Teacher Model           & \cellcolor[HTML]{fccece}{-}    & \cellcolor[HTML]{fccece}{-}   & \cellcolor[HTML]{fccece}{-}   & \cellcolor[HTML]{fccece}{-}   & \cellcolor[HTML]{fccece}{-}   & \cellcolor[HTML]{aac9f9}{single} & \cellcolor[HTML]{aac9f9}{single} & \cellcolor[HTML]{aac9f9}{single}  & \cellcolor[HTML]{aac9f9}{single} & \cellcolor[HTML]{aac9f9}{single}  & \cellcolor[HTML]{aac9f9}{single}  & \cellcolor[HTML]{c6f4c6}{ensemble} & \cellcolor[HTML]{c6f4c6}{ensemble} \\ \shline

DSA          & \cellcolor[HTML]{fccece}{No}  & \cellcolor[HTML]{c6f4c6}{Yes}   & \cellcolor[HTML]{c6f4c6}{Yes}   & \cellcolor[HTML]{c6f4c6}{Yes}  & \cellcolor[HTML]{c6f4c6}{Yes} & \cellcolor[HTML]{c6f4c6}{Yes} & \cellcolor[HTML]{fccece}{No} & \cellcolor[HTML]{fccece}{No}  & \cellcolor[HTML]{fccece}{No} & \cellcolor[HTML]{fccece}{No}  & \cellcolor[HTML]{fccece}{No}  &\cellcolor[HTML]{fccece}{No} & \cellcolor[HTML]{fccece}{No} \\
ZCA          & \cellcolor[HTML]{fccece}{No}  & \cellcolor[HTML]{fccece}{No}   & \cellcolor[HTML]{fccece}{No}   & \cellcolor[HTML]{c6f4c6}{Yes}  & \cellcolor[HTML]{fccece}{No} & \cellcolor[HTML]{c6f4c6}{Yes} & \cellcolor[HTML]{fccece}{No} & \cellcolor[HTML]{fccece}{No}  & \cellcolor[HTML]{fccece}{No} & \cellcolor[HTML]{fccece}{No}  & \cellcolor[HTML]{fccece}{No}  &\cellcolor[HTML]{fccece}{No} & \cellcolor[HTML]{fccece}{No} \\
ResizeCrop   & \cellcolor[HTML]{fccece}{No}  & \cellcolor[HTML]{fccece}{No}   & \cellcolor[HTML]{fccece}{No}   & \cellcolor[HTML]{fccece}{No}  & \cellcolor[HTML]{fccece}{No} & \cellcolor[HTML]{fccece}{No} & \cellcolor[HTML]{c6f4c6}{Yes} & \cellcolor[HTML]{c6f4c6}{Yes}  & \cellcolor[HTML]{c6f4c6}{Yes} & \cellcolor[HTML]{c6f4c6}{Yes}  & \cellcolor[HTML]{c6f4c6}{Yes}  & \cellcolor[HTML]{c6f4c6}{Yes} & \cellcolor[HTML]{c6f4c6}{Yes} \\
CropRange          & \cellcolor[HTML]{fccece}{-}  & \cellcolor[HTML]{fccece}{-}   & \cellcolor[HTML]{fccece}{-}   & \cellcolor[HTML]{fccece}{-}  & \cellcolor[HTML]{fccece}{-} & \cellcolor[HTML]{fccece}{-} & \cellcolor[HTML]{c6f4c6}{0.08, 1.0} & \cellcolor[HTML]{aac9f9}{0.5, 1.0}  & \cellcolor[HTML]{c6f4c6}{0.08, 1.0} & \cellcolor[HTML]{c6f4c6}{0.08, 1.0}  & \cellcolor[HTML]{c6f4c6}{0.08, 1.0}  & \cellcolor[HTML]{aac9f9}{0.5, 1.0} & \cellcolor[HTML]{c6f4c6}{0.08, 1.0} \\
PatchShuffle          & \cellcolor[HTML]{fccece}{No}  & \cellcolor[HTML]{fccece}{No}   & \cellcolor[HTML]{fccece}{No}   & \cellcolor[HTML]{fccece}{No}  & \cellcolor[HTML]{fccece}{No} & \cellcolor[HTML]{fccece}{No} & \cellcolor[HTML]{fccece}{No} & \cellcolor[HTML]{c6f4c6}{Yes}  & \cellcolor[HTML]{fccece}{No} & \cellcolor[HTML]{fccece}{No}  & \cellcolor[HTML]{fccece}{No}  & \cellcolor[HTML]{c6f4c6}{Yes} & \cellcolor[HTML]{fccece}{No} \\
CutMix          & \cellcolor[HTML]{fccece}{No}  & \cellcolor[HTML]{fccece}{No}   & \cellcolor[HTML]{fccece}{No}   & \cellcolor[HTML]{fccece}{No}  & \cellcolor[HTML]{fccece}{No} & \cellcolor[HTML]{fccece}{No} & \cellcolor[HTML]{c6f4c6}{Yes} & \cellcolor[HTML]{c6f4c6}{Yes}  & \cellcolor[HTML]{c6f4c6}{Yes} & \cellcolor[HTML]{c6f4c6}{Yes}  & \cellcolor[HTML]{c6f4c6}{Yes}  & \cellcolor[HTML]{c6f4c6}{Yes} & \cellcolor[HTML]{c6f4c6}{Yes} \\
\end{tabular}
}
\caption{Evaluation configurations of various dataset distillation methods. We separate agent model training hyperparameters (top) and data augmentation (bottom). For each row, different colors highlight the differences in the evaluation setting.}
\label{tab_configs}
\end{table}

\section{Motivation}

\subsection{Overview of Unfairness}
The conventional approach to evaluating dataset distillation methods relies on measuring the \textbf{test accuracy} of an agent model trained on the distilled dataset\footnote{Our discussion focuses exclusively on image classification datasets, as these are most frequently used.}.
However, we have identified substantial unfairness in this evaluation paradigm stemming from highly inconsistent training configurations for the agent model.
Table~\ref{tab_configs} presents a comparative analysis of training parameters and data augmentation employed by various dataset distillation methods on the same target dataset.
We use different colors to highlight the differences in the current dataset distillation evaluation settings.
We believe that the performance evaluated without a unified and standardized benchmark is not reliable for a fair comparison. 
Among these inconsistencies, two critical factors significantly undermine the fairness of current evaluation protocols: label representation (including the corresponding loss function) and data augmentation techniques.

\begin{figure}
    \centering
    \begin{subfigure}{0.48\textwidth}
        \includegraphics[width=0.99\textwidth]{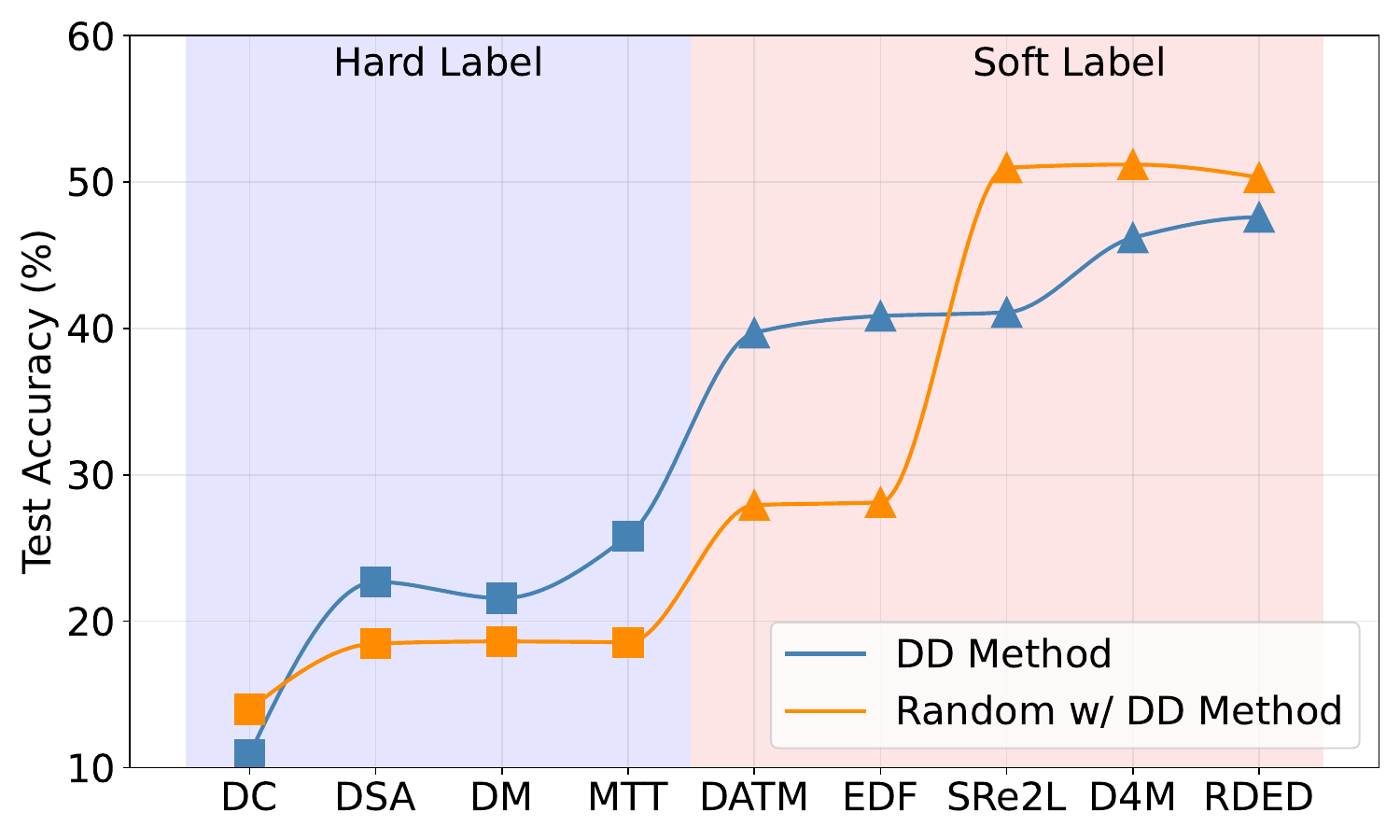}
        \caption{TinyImageNet IPC50}
    \end{subfigure}
    \hfill
    \begin{subfigure}{0.48\textwidth}
        \includegraphics[width=0.99\textwidth]{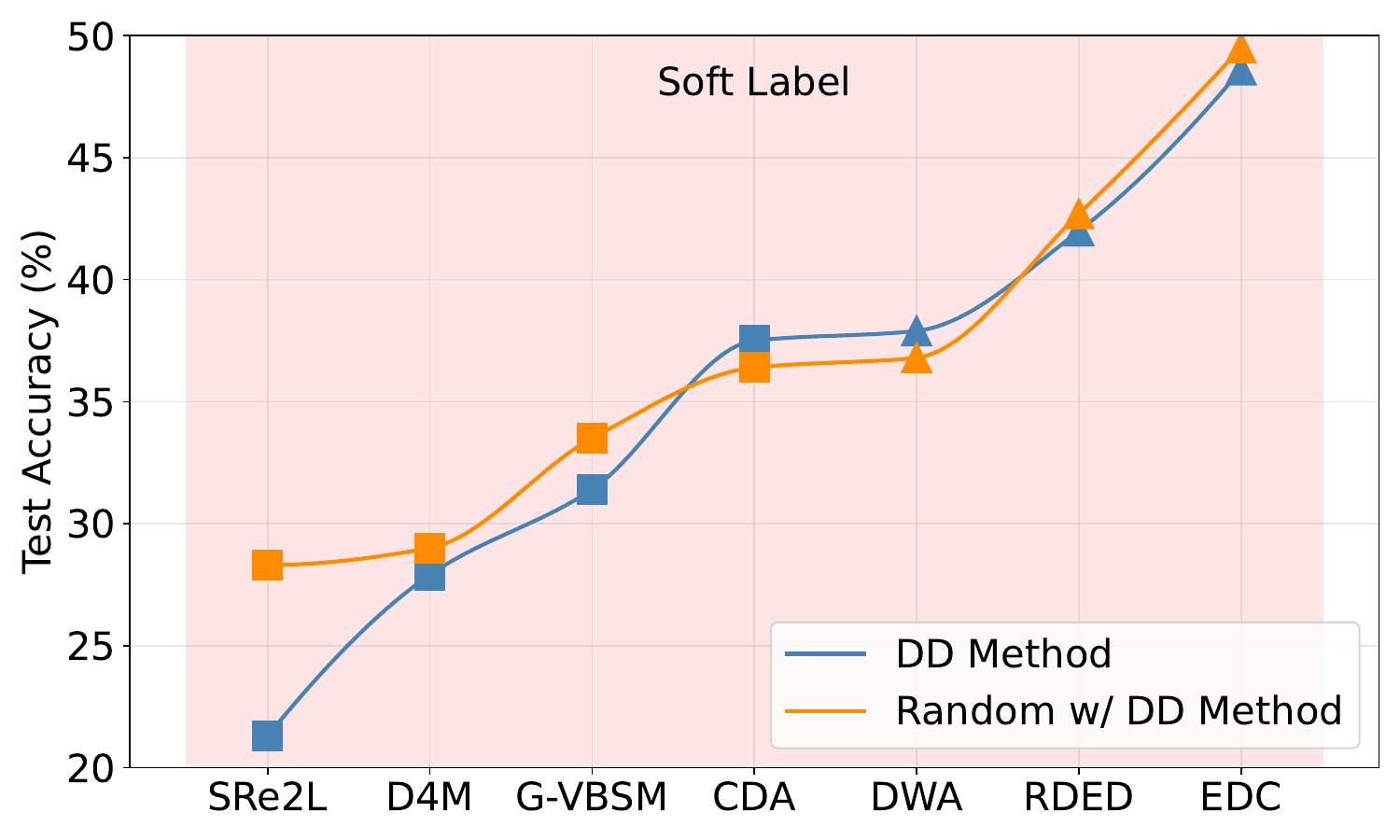}
        \caption{ImageNet1K IPC10}
    \end{subfigure}
    \caption{Test accuracies of the agent model trained on synthetic data distilled by various DD methods and on randomly selected data. Despite soft labels being able to significantly improve the test accuracy, DD methods may fail to outperform random selection under the same training setting.}
    \label{fig_motivation}
\end{figure}

\subsection{Soft Labels}
\paragraph{Soft labels significantly improve the test accuracy.}
Soft labels have been extensively employed in various domains, particularly in knowledge distillation tasks. 
Unlike hard labels, which assign discrete categorical values, soft labels represent probability distributions across class categories. 
These distributions are typically derived from the output logits of pretrained models. 
Recently, applying soft labels has emerged as a popular approach in evaluating dataset distillation methods. 
In this framework, each distilled image is associated with one or multiple soft labels generated by a pretrained teacher model. 
For instance, DATM~\citep{guo2023towards} concurrently optimizes synthetic data and corresponding soft labels during bi-level optimization procedures. 
SRe2L~\citep{yin2024squeeze} employs a teacher model to generate multiple soft labels per data sample at test time. 
Consequently, the training objective for an agent model on the distilled dataset becomes minimizing the loss (e.g., Kullback–Leibler divergence) between its output logits and these soft labels. 
Due to this knowledge distillation paradigm, evaluation metrics with soft labels consistently demonstrate substantially higher performance, as illustrated in Figure~\ref{fig_motivation}.

\begin{wraptable}{r}{0.45\textwidth}
    \centering
    \tablestyle{6pt}{1.2}
    \begin{tabular}{cl|cccc}
    \multicolumn{2}{c|}{dataset}  & \multicolumn{2}{c}{EDC} & \multicolumn{2}{c}{RDED} \\
    \multicolumn{2}{c|}{ipc}      & 10              & 50             & 10             & 50            \\ \shline
    \multicolumn{2}{c|}{w/ aug.}  &  48.6     &  58.0      &  42.0              &  56.5             \\
    \multicolumn{2}{c|}{w/o aug.} &  12.5     &  39.7      &   15.3             &   27.9           
    \end{tabular}
    \caption{Ablation on ImageNet1K. Data augmentation largely contributes to the high accuracy, especially on high-resolution datasets.}
    \label{tab_aug_ab}
\end{wraptable}
\paragraph{Improvements originate from knowledge distillation~\cite{qin2024label}, instead of synthetic data.}
We argue that the observed enhancement in test accuracy is predominantly attributable to knowledge distillation from soft labels, rather than any inherent improvement in the informativeness of the distilled data. 
To substantiate this claim, we conducted a comparative analysis examining test accuracies across random noises annotated with soft labels, randomly selected samples annotated with soft labels, and several baselines using soft labels. Throughout this experiment, we control all other training parameters the same across each baseline comparison, such as the identical teacher model, learning rate, and optimizer.

As demonstrated in Figure~\ref{fig_motivation}, data randomly selected from the original dataset but annotated with soft labels consistently outperforms baseline-distilled data in most cases. 
Moreover, even random noise patterns labeled with soft labels achieve non-negligible test accuracy, substantially exceeding random guessing. 
These findings conclude that while soft labeling techniques certainly elevate test accuracy metrics, they also obscure meaningful assessment of the intrinsic quality and representational capacity of the distilled data itself.

\subsection{Data Augmentation}
Data augmentation is a widely used technique to enhance model training performance. Current dataset distillation methods also apply various augmentation techniques during their evaluation process. As shown in Table~\ref{tab_configs}, there is significant diversity in the augmentation strategies used by existing dataset distillation methods, with different approaches typically adopting different sets of transformations. However, this variation makes it difficult to fairly evaluate and compare different dataset distillation methods because improvements in test accuracy brought about by data augmentation do not necessarily reflect the inherent quality of the distilled data itself.

To better demonstrate this claim, we conducted a comparative analysis of two established baseline methods, measuring their performance both with and without their respective data augmentation. As depicted in Table~\ref{tab_aug_ab}, a substantial portion of the reported performance gains can be directly attributed to augmentation rather than to the intrinsic quality of the distilled datasets. Therefore, similar to soft labels, these results highlight the need for new evaluation metrics that more accurately capture the true informational value of distilled data, instead of relying solely on raw test accuracy that can be inflated by augmentation techniques.

\section{DD-Ranking}

\subsection{Overview}
Motivated by the unfairness above, we introduce DD-Ranking.
DD-Ranking is an integrated and easy-to-use evaluation benchmark for dataset distillation (DD). 
It aims to provide a fair evaluation scheme for DD methods that can decouple the impacts from knowledge distillation and data augmentation to reflect the real informativeness of the distilled data.
Under the finding that the test accuracy no longer fits the need for fair and comprehensive evaluation, we design new metrics for both the label representation and data augmentation.

\subsection{Label-Robust Score}
\paragraph{Hard label recovery.} 
The initial goal of dataset distillation is to synthesize a small number of data points that do not need to come from the correct data distribution, but will, when given to the learning algorithm as training data, approximate the model trained on the original data~\citep{wang2018dataset}.
Given that almost all existing classification datasets use hard label annotation, we think it is crucial for DD methods to maintain good performance with hard labels.
To this end, we propose the \textbf{hard label recovery (HLR)}.
Specifically, for both hard-label-based and soft-label-based methods, we evaluate the test accuracy of the distilled data and that of the original dataset with hard labeling, denoted as $\text{acc}_\text{syn-hard}$ and $\text{acc}_\text{real-hard}$, respectively.
The hard label recovery is computed by taking the difference:
\begin{equation}
    \text{HLR} = \text{acc}_\text{real-hard} - \text{acc}_\text{syn-hard}
\end{equation}
A smaller HLR indicates that the distilled data enables the agent model to recover more of the performance of the same model trained on the full dataset.

\paragraph{Improvement over random.}
Despite the popularity of applying soft labels to evaluate DD methods, it's not fair to directly compare methods with soft labels against methods with hard labels.
Also, there isn't a unified recipe for soft-label-based training, and differences such as how many soft labels per sample, loss function, and temperature could significantly impact the results. 
This makes it difficult to compare different soft-label-based methods.
Thus, to make different methods comparable under mixed label types, we propose \textbf{improvement over random (IOR)}.
This metric is based on the common sense that any DD method should at least outperform random selection under the same training recipe, and we use the relative performance improvements over random selection to compare any pair of DD methods.
Specifically, denote the test accuracy of the model trained on synthetic data with any label type and that on a randomly selected subset (the capacity (e.g., image per class) is kept the same as the synthetic data) with that label type as $\text{acc}_{\text{syn-any}}$ and $\text{acc}_{\text{rdm-any}}$, respectively.
For each DD method, we keep all of its evaluation settings (such as data augmentation, loss function, learning rate, etc.) unchanged when training the agent model on random data.
Then, the IOR is computed by:
\begin{equation}
    \text{IOR} = \text{acc}_{\text{syn-any}} - \text{acc}_{\text{rdm-any}}
\end{equation}
IOR is positively related to the performance of DD methods.
By doing so, we can effectively disentangle the improvement brought solely by knowledge distillation and reflect the true informativeness of the distilled data.

\paragraph{Label-robust score}
Combining hard label recovery (HLR) and improvement over random (IOR), we present the label-robust score (LRS).
LRS first takes a weighted sum $\alpha$ of HLR and IOR via a weight parameter $\lambda$ as follows:
\begin{equation}
    \alpha = \lambda \text{IOR} - (1 - \lambda) \text{HLR}
\end{equation}
We assign a negative mark to HLR so that both parts of the sum are positively correlated with the performance.
The raw range of $\alpha$ is between $[-1, 1]$, so we normalize LRS to the range $[0, 1]$ by letting $\text{LRS} = 100\% \times (e^{\alpha} - e^{-1}) / (e - e^{-1})$.
A higher LRS indicates that the distilled dataset of the corresponding method is more robust to the label representation and has richer information.

\subsection{Augmentation-Robust Score}
\label{sec_ars}
Data augmentation, as a trick to enhance model training, doesn't reveal the quality of the dataset itself.
Thus, the improvement in test accuracy brought merely by data augmentation at test time should not be attributed to the effectiveness of the dataset distillation method.
To disentangle data augmentation's impact, we introduce the \textbf{augmentation-robust score (ARS)} which continues to leverage the relative improvement over randomly selected data.
Specifically, we first evaluate synthetic data and a randomly selected subset under the same setting to obtain $\text{acc}_\text{syn-aug}$ and $\text{acc}_\text{rdm-aug}$ (same as $\text{IOR}$).
Next, we evaluate both synthetic data and random data again without the data augmentation, and results are denoted as $\text{acc}_\text{syn-naug}$ and $\text{acc}_\text{rdm-naug}$.

We claim that an informative subset via distillation should surpass any randomly selected subset of the same size, regardless of the use of data augmentation.
Thus, both differences, $\text{acc}_\text{syn-aug} - \text{acc}_\text{rdm-aug}$ and $\text{acc}_\text{syn-naug} - \text{acc}_\text{rdm-naug}$, are positively correlated to the real informativeness of the distilled dataset.
We take a weighted sum of the two differences
\begin{equation}
    \beta = \gamma (\text{acc}_\text{syn-aug} - \text{acc}_\text{rdm-aug}) + (1 - \gamma) (\text{acc}_\text{syn-naug} - \text{acc}_\text{rdm-naug})
\end{equation}
and use a similar normalization method to compute $\text{ARS}$. A higher ARS indicates that the distilled dataset of the corresponding method is more robust to data augmentation.




\section{Results}

\subsection{Evaluation Settings}

\paragraph{Baseline.}
We evaluate a wide range of representative works in dataset distillation.
For hard-label methods, we evaluate DC~\citep{zhao2021dataset}, DSA~\citep{zhao2021dsa}, MTT~\citep{cazenavette2022dataset}, DM~\citep{zhao2021datasetdm}, and DataDAM~\citep{sajedi2023datadam}.
For soft-label methods, we evaluate SRe2L~\citep{yin2024squeeze}, DATM~\citep{guo2023towards}, EDF~\citep{wang2025emphasizingdiscriminativefeaturesdataset}, DWA~\citep{du2024diversitydriven}, RDED~\citep{sun2024diversity}, CDA~\citep{yin2024dataset}, EDC~\citep{shao2024elucidating}, and G-VBSM~\citep{shao2024generalizedlargescaledatacondensatio}.
In the case where the method provides its distilled data, we adopt it directly.
In the case where the distilled data is absent, we strictly follow their implementation provided in both the paper and code repo to replicate their results.

\paragraph{Dataset.}
We report DD-Ranking benchmarking results on the four \textbf{existing datasets}: CIFAR-10~\citep{Krizhevsky2009LearningML}, CIFAR-100~\citep{Krizhevsky2009LearningML}, TinyImageNet~\citep{Le2015TinyIV}, and ImageNet1K~\citep{russakovsky2015imagenetlargescalevisual}.
The resolution of images in CIFAR-10 and CIFAR-100 is $32\times 32$.
The resolution of images in TinyImageNet is $64\times 64$.
The resolution of images in ImageNet1K is $224\times 224$.
We only report ARS results on ImageNet1K due to space limit. More results can be found in our leaderboard.

\paragraph{Model.}
For each baseline method, we use the model architecture reported in the paper for evaluation. This includes ConvNet of depth 3 and 4 with instance normalization, ConvNet of depth 3 and 4 with batch normalization, and ResNet-18~\citep{He2015DeepRL}.
Additionally, to validate the robustness of DD-Ranking on different model architectures, we incorporate AlexNet~\citep{AlexNet}, ResNet-50, VGG-11~\citep{VGG11}, Swin-Transformer-tiny~\citep{Swin-T}, and Vision-Transformer-base~\citep{ViT}.

\paragraph{DD-Ranking evaluation.}
The evaluation is performed \textbf{5 times} with different random seeds. We report \textbf{the mean value} in the following tables. Standard deviations are reported in the appendix.
When computing the accuracy under hard labels, we perform the hyperparameter search for the learning rate and report the best one.
When computing the accuracy under soft labels, we regard the learning provided by each method as the \textbf{optimal learning rate} by default, and the learning rate search is performed for random selection.

\subsection{Label-Robust Score}

\begin{table}[t]
\centering
\tablestyle{8pt}{1.3}
\begin{tabular}{c|cccccccccc}
ipc        & \multicolumn{3}{c}{1} & \multicolumn{3}{c}{10} & \multicolumn{3}{c}{50} \\
metric        & HLR$\downarrow$   & IOR$\uparrow$   & LRS$\uparrow$   & HLR$\downarrow$   & IOR$\uparrow$   & LRS$\uparrow$   & HLR$\downarrow$   & IOR$\uparrow$   & LRS$\uparrow$   \\ \shline \rowcolor{hardlabel}
DC         &  52.7     &  12.4     &  19.1     & 36.7    &  18.5     & 23.2      &  26.3      &  12.3     &  24.0     \\ \rowcolor{hardlabel}
DSA        &  58.9     &  13.2     & 18.2      & 35.1   & 19.6      & 23.7      &  27.4    & 11.0      & 23.5      \\ \rowcolor{hardlabel}
MTT        &  42.2     &  27.6     & 23.9      & \textbf{23.7}   & 30.9      & 28.4      &  \textbf{16.5}    &  20.5     & 27.8      \\ \rowcolor{hardlabel}
DM         &  61.4     &  8.7     &  17.0     &  39.4   & 16.1      & 22.2      & 25.1       &  12.7     & 24.3      \\ \rowcolor{hardlabel}
DATADAM    &  49.9     &  15.6     & 20.0      &  34.8  & 19.9      & 23.8      & 21.9     & 15.8      &  25.6     \\ \shline \rowcolor{softlabel}
DATM       &  \textbf{41.9}     &  \textbf{30.8}     & \textbf{24.6}      & 26.8   & \textbf{35.1}   &  \textbf{28.7}     &  18.9      & \textbf{23.9}      &  \textbf{28.0}     \\ \rowcolor{softlabel}
SRe2L      &  69.9     &  -0.3     & 14.3      & 67.8   & -5.7   &  13.8     &  62.9    & -6.5    & 14.4      \\ \rowcolor{softlabel}
RDED       &  60.6     &  2.4     & 16.2      &  50.7   & 1.1    &  17.6     &  36.0      & -1.6      & 19.6      \\ \rowcolor{softlabel}
D4M        &  51.1     &  6.7     &  18.4     & 39.9    & 9.1    &  20.8     & 27.0    &  6.6     &  22.8    
\end{tabular}
\caption{Label-robust score evaluation results on CIFAR-10. We also report the hard-label recovery and improvement over random for a more comprehensive comparison. The color scheme corresponds to that of Figure~\ref{fig_motivation}. The $\lambda$ is set to 0.5 for this and the following results. On CIFAR-10, hard-label-based methods perform generally better.}
\label{tab_cifar10_main}
\end{table}
\begin{table}[t]
\centering
\tablestyle{8pt}{1.3}
\begin{tabular}{c|ccccccccc}
ipc           & \multicolumn{3}{c}{1} & \multicolumn{3}{c}{10} & \multicolumn{3}{c}{50} \\
metric        & HLR$\downarrow$   & IOR$\uparrow$   & LRS$\uparrow$   & HLR$\downarrow$   & IOR$\uparrow$   & LRS$\uparrow$   & HLR$\downarrow$   & IOR$\uparrow$   & LRS$\uparrow$   \\ \shline \rowcolor{hardlabel}
DC   &  39.4  &  8.4     &  20.8     &  25.5      & 12.7      & 24.2      &  21.8    & 1.1      &  22.7     \\ \rowcolor{hardlabel}
DSA  & 46.0   &  8.5     &  19.6     &  26.1      & 13.5      & 24.3      &  21.2    & 2.0      & 23.0      \\ \rowcolor{hardlabel}
MTT  & 35.2   &  16.7    &  23.1     &  \textbf{18.0}      & \textbf{20.7}      & \textbf{27.5}      &  12.1    & 11.6      &  26.8     \\ \rowcolor{hardlabel}
DM   & 48.0   &  6.1     &  18.9     &  30.1      & 10.7      & 23.0      &  16.6    & 7.2      & 24.9      \\ \rowcolor{hardlabel}
DATADAM  &  45.2   & 9.1   &  19.9    &  25.9      & 14.8      & 24.6     &  12.4    & 11.8      &  26.8     \\ \shline \rowcolor{softlabel}
DATM  &  \textbf{24.1}    &  \textbf{18.5}  & \textbf{25.7}      & 18.9    & 18.4    &  26.8     & \textbf{10.3}    & \textbf{26.1}     &  \textbf{30.4}     \\ \rowcolor{softlabel}
SRe2L  &  52.7     & -1.9    &  16.7     &  50.5    & -14.8      &  15.0     &  46.2      & -11.5      &  16.2     \\ \rowcolor{softlabel}
RDED  & 45.6   &  -0.5     & 18.1      &  37.5      &  -1.2     &  19.4     &  27.3  & -1.5  & 21.2      \\ \rowcolor{softlabel}
D4M  & 30.9      & 10.0    & 22.7      &  40.1      &  9.7     &  20.9     &  26.7  &  13.5     & 24.2
\end{tabular}
\caption{LRS, HLR, and IOR evaluation results on CIFAR-100. DATM constantly performs the best and outperforms random selection to a large extent. This implies that soft labels are effective in improving synthetic data when used properly.}
\label{tab_cifar100_main}
\end{table}
\begin{table}[t]
\centering
\tablestyle{8pt}{1.3}
\begin{tabular}{c|ccccccccc}
ipc           & \multicolumn{3}{c}{1} & \multicolumn{3}{c}{10} & \multicolumn{3}{c}{50} \\
metric        & HLR$\downarrow$   & IOR$\uparrow$   & LRS$\uparrow$   & HLR$\downarrow$   & IOR$\uparrow$   & LRS$\uparrow$   & HLR$\downarrow$   & IOR$\uparrow$   & LRS$\uparrow$   \\ \shline \rowcolor{hardlabel}
DC    &  28.6     &  3.9     &  22.0     &  21.5      &  7.1     &  23.9     &  21.3      &  -2.1    &  22.2     \\ \rowcolor{hardlabel}
DSA   &  30.3     &  3.7     &  21.6     &  20.3      &  6.8     &  24.1     &  17.6      &  7.6     &  25.8     \\ \rowcolor{hardlabel}
MTT   &  30.7     &  5.8     &  21.9     &  \textbf{15.6}      &  14.6     &  \textbf{26.7}     &  15.6      & 10.2      &  26.4     \\ \rowcolor{hardlabel}
DM    &  36.7     &  2.3     &  20.2     &  26.2      &  7.5     &  23.1     &  18.9      &  5.3     &  24.1     \\ \shline \rowcolor{softlabel}
DATM  & \textbf{25.4}      &  8.6     &  \textbf{23.5}     &  18.3      &  14.2     &  26.0     &   \textbf{13.5}     &  15.1     &  27.2     \\ \rowcolor{softlabel}
EDF  &  25.8     &  \textbf{9.2}     &  \textbf{23.5}     & 18.5       &  \textbf{15.4}     &  26.2     &  13.8      & 15.9      &  \textbf{27.3}     \\ \rowcolor{softlabel}
SRe2L &  45.6     &  -1.8    &  15.4     &  43.6      &  -8.5     &  17.1     &  33.6      &  -9.6     &  18.6     \\ \rowcolor{softlabel}
RDED  &  34.0     &  3.9     &  21.0     &  25.6      &  1.8     &  23.7     &  15.2      &  -0.6     &  23.7     \\ \rowcolor{softlabel}
D4M   &   40.6   &  -3.0   & 18.6      &  35.6      &  -5.8     &  18.9     &   27.7     &  12.8     & 23.8
\end{tabular}
\caption{LRS, HLR, and IOR evaluation results on TinyImageNet. For decoupled methods, D4M appears to be more effective when IPC is large, and RDED performs better at smaller IPCs.}
\label{tab_tiny_main}
\end{table}
\begin{table}[t]
\centering
\tablestyle{8pt}{1.3}
\begin{tabular}{c|ccccccccc}
ipc           & \multicolumn{3}{c}{1} & \multicolumn{3}{c}{10} & \multicolumn{3}{c}{50} \\
metric        & HLR$\downarrow$   & IOR$\uparrow$   & LRS$\uparrow$   & HLR$\downarrow$   & IOR$\uparrow$   & LRS$\uparrow$   & HLR$\downarrow$   & IOR$\uparrow$   & LRS$\uparrow$   \\ \shline \rowcolor{softlabel}
SRe2L    &  56.3     &  -1.5     &  16.2     &  55.0      &  -15.6     &  14.2     &  53.4      &  -13.2     & 14.8  \\ \rowcolor{softlabel}
RDED     &  \textbf{55.7}     &  \textbf{1.6}     &  \textbf{16.8}     &  \textbf{50.2}      &  -0.6     &  \textbf{17.4}     &  \textbf{39.8}      &  -3.6     & \textbf{22.9}  \\ \rowcolor{softlabel}
D4M      &  55.9     &  -0.6     &  15.6     &  53.0      & -7.7  & 15.8      &  43.7      &  -5.8     &  17.6 \\ \rowcolor{softlabel}
DWA      &  56.1     &  -1.2     &  16.3     &  54.4      &  -4.1     &    16.1   &  49.7      &  -7.8     &  16.3 \\ \rowcolor{softlabel}
CDA      &  56.2     &  -2.5     &  16.1     &  54.9      &  -8.6     &     15.3  &   52.0     &   -6.7    &  16.1 \\ \rowcolor{softlabel}
EDC      &   55.7    &   -0.8    &  16.4     &    52.0    &    \textbf{-0.4}   &    17.1   &    41.3    &   \textbf{-0.1}    & 18.9  \\ \rowcolor{softlabel}
G-VBSM   &   56.3    &    -1.2   &  16.3     &    55.0    &    -7.3   &     15.5  &    44.9    &   -5.9    &  17.4 \\ 
\end{tabular}
\caption{LRS, HLR, and IOR evaluation results on ImageNet1K. Notably, existing DD methods (mainly decoupled) hardly outperform random selection and perform, and fail to perform well when switched to hard labels.}
\label{tab_in1k_main}
\end{table}

\paragraph{Results on CIFAR-10, CIFAR-100, and TinyImageNet.}
Tables~\ref{tab_cifar10_main}, \ref{tab_cifar100_main}, and \ref{tab_tiny_main} present LRS evaluation results on CIFAR-10, CIFAR-100, and TinyImageNet, respectively. 
Among hard-label-based methods, trajectory matching (MTT) achieves the best performance, outperforming both gradient matching approaches (DC and DSA) and distribution matching methods (DM and DataDAM).
As IPC increases, the distribution matching methods perform better than the gradient matching methods.
Within the soft-label-based category, methods that optimize one-to-one soft labels jointly with synthetic data (DATM) demonstrate superior performance compared to approaches that directly utilize multiple soft labels from teacher models (D4M, SRe2L, and RDED). 
D4M, which employs a generative modeling approach, outperforms decoupled methods, especially when IPC increases.
Across all methods, DATM emerges as the strongest baseline. Notably, hard-label-based methods yield results closer to full-dataset performance with hard labels and exhibit greater improvement over random data selection compared to their soft-label counterparts.

\paragraph{Results on ImageNet1K.}
Table~\ref{tab_in1k_main} presents LRS results of various methods on ImageNet1K. All existing methods capable of efficiently scaling to ImageNet1K employ soft labeling techniques. Remarkably, current DD methods consistently underperform random selection across most IPC settings when soft labeling is also applied to randomly selected data. 
This performance gap widens as IPC increases. While these methods achieve notably high accuracy when using soft labels, their performance under hard labels deteriorates significantly, revealing a substantial gap compared to the real dataset.

\paragraph{Findings.}
Based on these results, we identify three key insights.

\textit{i) Test accuracy is not a reliable metric when soft labels are employed.}
Soft labels demonstrate even higher effectiveness on random data. 
Notably, on TinyImageNet and ImageNet1K, classifiers trained on random data with soft labels consistently \textbf{outperform} those trained on DD-synthesized data. 
While DATM maintains an advantage over random selection on TinyImageNet, this improvement diminishes substantially when soft labels are applied to random data. 
This observation reinforces our claim that accuracy improvements with soft labels primarily stem from knowledge distillation rather than the intrinsic informativeness of synthetic data.

\textit{ii) Soft labels enhance synthetic dataset informativeness when jointly optimized.}
Among soft-label-based methods, DATM and EDF employ a distinct approach by assigning unique soft labels to each sample and jointly optimizing both samples and labels during distillation. 
Unlike generative and decoupled methods that generate soft labels at test time, these optimized soft labels improve synthetic data quality, as evidenced by superior LRS scores.
This demonstrates that integrating soft labels into the training process can meaningfully enhance synthetic data quality.

\textit{iii) Matching-based methods remain the strongest baselines.}
Despite computational limitations that restrict their scalability to large-scale datasets like ImageNet1K, matching-based methods (encompassing gradient, trajectory, and feature matching) consistently produce more effective distilled datasets. Besides, RDED and D4M appear to be more effective among decoupled methods, implying the importance of the realism of synthetic data.

\subsection{Augmentation-Robust Score}

\begin{table}[t]
\centering
\tablestyle{2pt}{1.2}
\resizebox{\textwidth}{!}{%
\begin{tabular}{c|ccccccccc}
ipc           & \multicolumn{3}{c}{1} & \multicolumn{3}{c}{10} & \multicolumn{3}{c}{50} \\
metric        & IOR w/o aug$\uparrow$   & IOR w/ aug $\uparrow$   & ARS$\uparrow$   & IOR w/o aug$\uparrow$   & IOR w/ aug $\uparrow$   & ARS$\uparrow$   & IOR w/o aug$\uparrow$   & IOR w/ aug $\uparrow$   & ARS$\uparrow$   \\ \shline \rowcolor{softlabel}
SRe2L    &  -1.2     &  -1.5     &  26.3     &  -4.4      &  -15.6     &  22.9     &  -21.0      &  -13.2     & 20.2  \\ \rowcolor{softlabel}
RDED     &  0.8    &  \textbf{1.6}     &  27.4     &  5.6      &  -0.6     &  28.1     &  2.0     &  -3.6     & 26.7  \\ \rowcolor{softlabel}
D4M      &  -0.3     &  -0.6     &  26.7     &  -0.5      & -7.7  & 25.2      &  -2.0      &  -5.8     &  25.3 \\ \rowcolor{softlabel}
DWA      &  -1.2     &  -1.2     &  26.4     &  -4.0      &  -4.1     &    25.2  &  -13.0      &  -7.8     &  22.7 \\ \rowcolor{softlabel}
CDA      &  -1.1     &  -2.5     &  26.1     &  -4.9      &  -8.6     &    24.1  &   -14.1     &   -6.7    &  22.7 \\ \rowcolor{softlabel}
EDC      &   -0.5    &   -0.8    &  26.6     &    -0.3    &    \textbf{-0.4}   &    26.8   &    -3.2    &   \textbf{-0.1}    & 26.2  \\ \rowcolor{softlabel}
G-VBSM   &   -1.2    &    -1.2   &  26.4     &    -7.9    &    -7.3   &   23.8  &    -18.0    &   -5.9    &  22.1 \\ 
\end{tabular}
}

\caption{Augmentation-robust score (ARS) evaluation results on ImageNet1K. We report both IOR w/ aug ($\text{acc}_\text{syn-aug} - \text{acc}_\text{rdm-aug}$) and IOR w/o aug ($\text{acc}_\text{syn-naug} - \text{acc}_\text{rdm-naug}$). $\gamma$ is 0.5 by default.}
\label{tab_in1k_ars}
\end{table}
Table~\ref{tab_in1k_ars} presents ARS performance metrics for various DD methods applied to ImageNet1K, including IOR results with and without data augmentation as introduced in Section~\ref{sec_ars}. Most existing decoupled and generative DD methods fail to surpass random selection regardless of augmentation status. Without data augmentation, the performance disparity between DD methods and random selection widens as IPC increases. These findings demonstrate that contemporary DD approaches, despite their heavy reliance on data augmentation strategies, frequently underperform when these same augmentation techniques are applied to simple random selection. Notably, when augmentation is excluded from evaluation, the performance gap between certain DD methodologies and random selection becomes more pronounced, further supporting our assertion that conventional test accuracy metrics no longer serve as an equitable evaluation criterion in this domain.

\subsection{Analysis}

\paragraph{Robust to model architecture.}
\begin{figure}[t]
    \centering
    \includegraphics[width=1.0\textwidth]{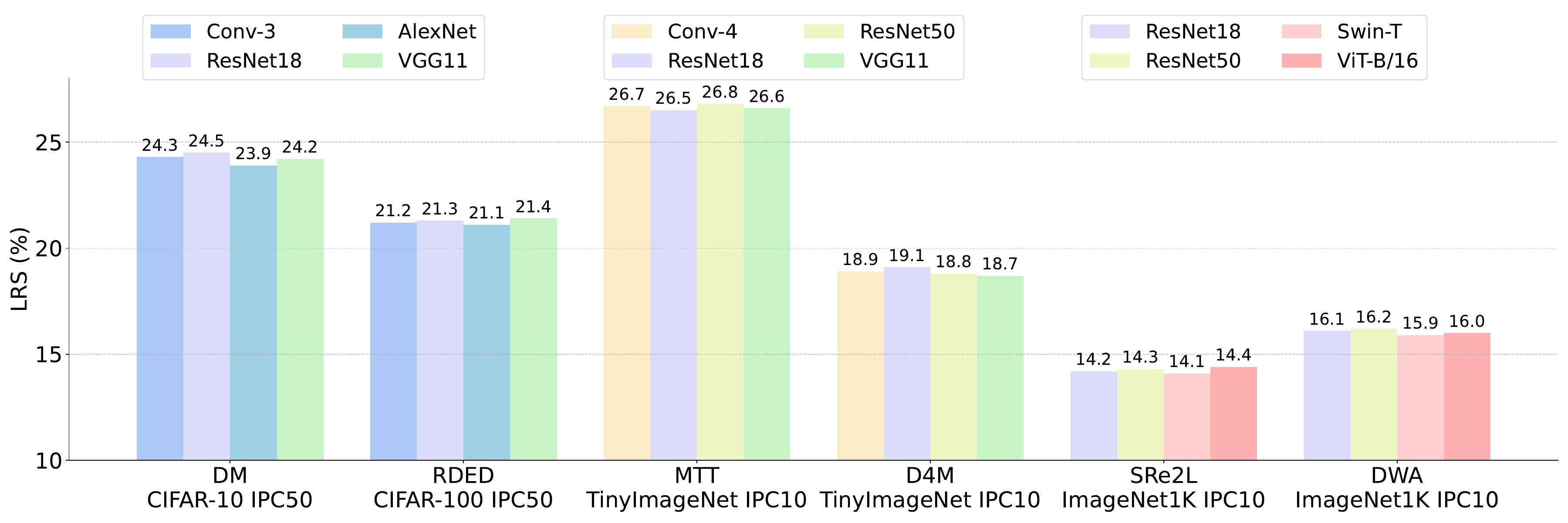}
    \caption{Label-robust scores of various methods with four different agent model architectures. The LRS fluctuation is minimal for each method, indicating that DD-Ranking is robust to different model architectures.}
    \label{fig_robust_arch}
\end{figure}
Cross-architecture evaluation is an important experiment for dataset distillation methods. Specifically,  different models architectures are used to evaluate the synthetic data. Despite variations in raw test accuracy across model architectures, the metric used to evaluate dataset distillation performance should remain consistent, with minimal fluctuation in metric values. As shown in Figure~\ref{fig_robust_arch}, the LRS results for six methods across different settings, each tested with four distinct model architectures, demonstrate high consistency. This consistency validates the robustness of our benchmark across different model architectures.

\paragraph{Robust to soft labels.}
\begin{figure}[t]
    \centering
    \includegraphics[width=1.0\textwidth]{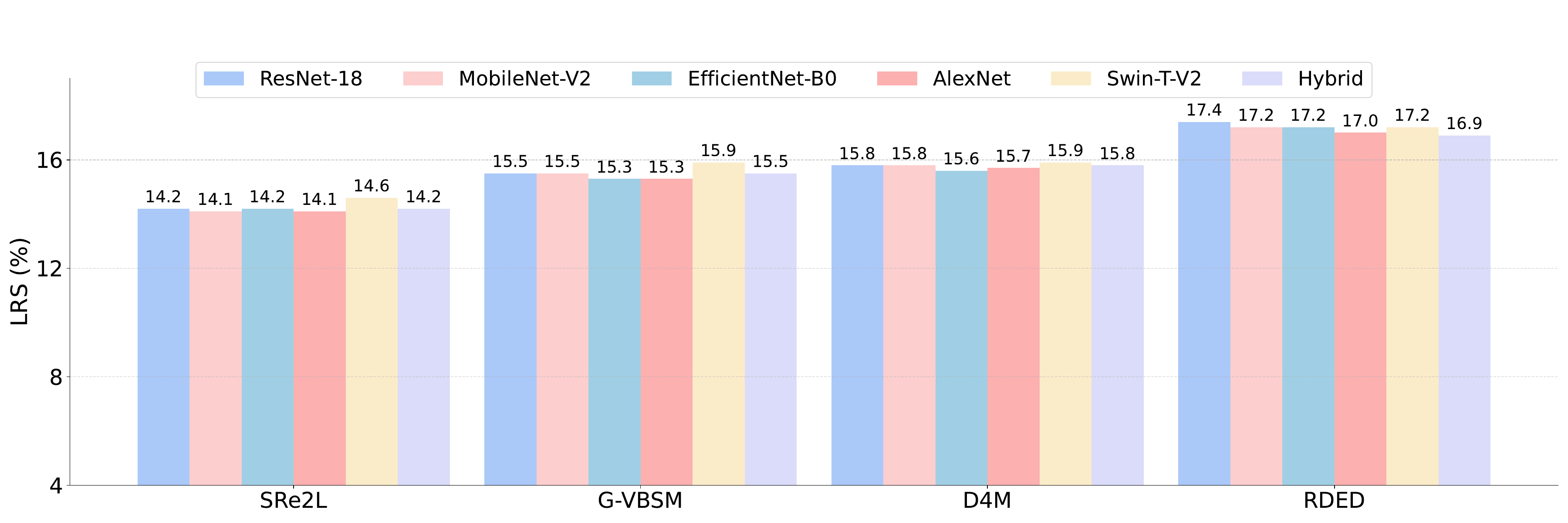}
    \caption{Label-robust scores of decoupled distillation methods with different teacher model architectures. The LRS fluctuation is minimal for each method, indicating that DD-Ranking is robust to soft labels generated by different models.}
    \label{fig_robust_label}
\end{figure}
In decoupled dataset distillation~\cite{yin2024squeeze, shao2024generalized, su2024d4m, sun2024diversity}, epoch-wise soft labels constitute a crucial component of the synthetic dataset. Recent studies~\cite{shao2024generalized,shao2024elucidating,cui2025dataset} have explored improving test accuracy by leveraging stronger teacher models to provide soft labels without altering the synthetic data itself. However, the validity of this technique remains insufficiently investigated. As shown in Figure~\ref{fig_robust_label}, whether through the use of different teacher models or advanced hybrid soft label strategies by fusing soft labels generated by multiple teachers, our proposed LRS consistently exhibits strong robustness, thereby validating its reliability across diverse soft label settings.

\section{Related Works}

\paragraph{Hard-label-based dataset distillation methods.}
Hard-label-based DD methods assign categorical labels to synthetic samples, the same as the labels of the real dataset.
Matching-based methods are known as representative hard-label-based methods. \textit{i) Gradient matching}: Synthetic data is optimized such that the gradients they induce on a neural network closely match those from real data. Following the pioneering work of Dataset Condensation (DC)~\cite{zhao2021dataset}, various works have improved gradient matching, such as DSA~\citep{zhao2021dsa}, DCC~\citep{DCC}, and LCMat~\citep{LCMat}.  \textit{ii) Trajectory matching}: Synthetic data are optimized by aligning the training dynamics of models trained on synthetic data with those trained on real data. MTT~\cite{cazenavette2022dataset} first introduced this approach, where synthetic data is optimized by aligning the training dynamics of models trained on synthetic data with those trained on real data. Building on this, various methods such as TESLA~\cite{cui2023scaling}, FTD~\citep{du2023minimizing}, and ATT~\citep{ATT} further enhance trajectory matching by improving memory efficiency and reducing trajectory errors. \textit{iii) Feature matching} is an alternative to gradient or trajectory-based distillation, where synthetic data is optimized to induce similar internal representations as real data. Represented by CAFE~\cite{wang2022cafe}, DM~\cite{zhao2023dataset}, and DataDAM~\cite{sajedi2023datadam}, this approach offers a lightweight framework with comparable performances, especially on large IPC settings.

\paragraph{Soft-label-based dataset distillation methods}.
DD methods using soft labels employ knowledge distillation during evaluation.
Each synthetic sample is assigned to one or multiple soft labels generated by a pretrained teacher model.
Among matching-based methods, DATM~\citep{guo2023towards}, PAD~\citep{li2024prioritizealignmentdatasetdistillation}, and EDF~\citep{wang2025emphasizingdiscriminativefeaturesdataset} optimize the soft labels jointly with synthetic data during trajectory matching.
Recently, decoupled methods have demonstrated strong scalability on large datasets such as ImageNet1K by decoupling the bi-level optimization.
SRe2L~\citep{yin2024squeeze} first proposed a three-stage "squeeze, recover, and relabel" paradigm.
During the relabel stage, soft labels are generated and saved for each synthetic sample. 
Following this approach, CDA~\citep{yin2024dataset}, DWA~\citep{sun2024diversity}, EDC~\citep{shao2024elucidating}, and G-VBSM~\citep{shao2024generalizedlargescaledatacondensatio} further improve the performance from both data and soft label perspectives.
RDED synthesizes condensed data by concatenating core image patches.
D4M employs diffusion models to generate high-quality synthetic images.

\paragraph{Dataset distillation benchmark.}
A notable challenge for dataset distillation is the lack of comprehensive benchmarks.
DC-Bench~\citep{cui2022dc} is the first large-scale standardized benchmark for dataset condensation methods in general.
It provides a comprehensive evaluation for several dataset distillation methods and coreset selection methods.
Comp-DD is proposed in EDF~\citep{wang2025emphasizingdiscriminativefeaturesdataset} targeting dataset distillation in complex scenarios.
It extracts new subsets from ImageNet1K based on the complexity metric.
However, both benchmarks no longer satisfy the need for fair evaluation of dataset distillation methods under the soft label trend.
Therefore, DD-Ranking is proposed to solve this problem.
\section{Conclusion and Future Work}
We propose DD-Ranking, a new benchmark that provides a fair and comprehensive evaluation for dataset distillation.
DD-Ranking is well motivated by the unfairness originated from inconsistent training settings of existing DD evaluation, especially the use of soft labels and data augmentation.
To this end, DD-Ranking introduces both label robust score and augmentation robust score to disentangle the effect of knowledge distillation via soft labeling and data augmentation, and ultimately reveal the true informativeness of distilled datasets.
Hopefully, DD-Ranking can facilitate the development of dataset distillation towards improving data quality instead of accuracy.
DD-Ranking is already open-source as a PyPI package with detailed documentation.
One potential limitation of the current DD-Ranking is that we only support methods for image classification dataset distillation.
We are aware that several works~\citep{zhou2023datasetquantization, wu2024visionlanguagedatasetdistillation} have extended dataset distillation to other tasks and modalities.
In the future, we will constantly integrate more baseline methods into our benchmark and extend DD-Ranking to other modalities.

\bibliographystyle{plain} 
\bibliography{neuips_2025}

\newpage
\appendix
\section{DD-Ranking Team}
\label{dd_ranking_team}
We provide the full list of DD-Ranking team members as follows (* denotes equal contribution):

\begin{multicols}{2}
\begin{itemize}
\item Zekai Li* (National University of Singapore)
\item Xinhao Zhong* (National University of Singapore)
\item Samir Khaki (University of Toronto)
\item Zhiyuan Liang (National University of Singapore)
\item Yuhao Zhou (National University of Singapore)
\item Mingjia Shi (National University of Singapore)
\item Dongwen Tang (National University of Singapore)
\item Ziqiao Wang (National University of Singapore)
\item Wangbo Zhao (National University of Singapore)
\item Xuanlei Zhao (National University of Singapore)
\item Mengxuan Wu (National University of Singapore)
\item Haonan Wang (National University of Singapore)
\item Ziheng Qin (National University of Singapore)
\item Dai Liu (Technical University of Munich)
\item Kaipeng Zhang (Shanghai AI Lab)
\item Tianyi Zhou (A*STAR)
\item Zheng Zhu (Tsinghua University)
\item Kun Wang (University of Science and Technology of China)
\item Shaobo Wang (Shanghai Jiao Tong University)
\item Guang Li (Hokkaido University)
\item Junhao Zhang (National University of Singapore)
\item Jiawei Liu (National University of Singapore)
\item Zhiheng Ma (SUAT)
\item Linfeng Zhang (Shanghai Jiao Tong University)
\item Yiran Huang (Technical University of Munich)
\item Lingjuan Lyu (Sony)
\item Jiancheng Lv (Sichuan University)
\item Yaochu Jin (Westlake University)
\item Zeynep Akata (Technical University of Munich)
\item Jindong Gu (Oxford University)
\item Peihao Wang (University of Texas at Austin)
\item Mike Shou (National University of Singapore)
\item Zhiwei Deng (Google DeepMind)
\item Qian Zheng (Zhejiang University)
\item Hao Ye (Xiaomi)
\item Shuo Wang (Baidu)
\item Xiaobo Wang (Chinese Academy of Science)
\item Yan Yan (University of Illinois at Chicago)
\item Yuzhang Shang (University of Illinois at Chicago)
\item George Cazenavette (Massachusetts Institute of Technology)
\item Xindi Wu (Princeton University)
\item Justin Cui (University of California, Los Angeles)
\item Tianlong Chen (University of North Carolina at Chapel Hill)
\item Angela Yao (National University of Singapore)
\item Baharan Mirzasoleiman (University of California, Los Angeles)
\item Hakan Bilen (University of Edinburgh)
\item Manolis Kellis (Massachusetts Institute of Technology)
\item Konstantinos N. Plataniotis (University of Toronto)
\item Bo Zhao (Shanghai Jiao Tong University)
\item Zhangyang Wang (University of Texas at Austin)
\item Yang You (National University of Singapore)
\item Kai Wang (National University of Singapore)
\end{itemize}
\end{multicols}

Zekai and Xinhao \textit{contribute equally} to this work.
Zekai serves as the \textit{project lead}, and Kai Wang is the \textit{corresponding author}.

\section{Additional Experiment Results}

Results from Table~\ref{tab_cifar10_main} to Table~\ref{tab_in1k_main} are computed by letting $\lambda=0.5$.
By default, we treat the hard label recovery and improvement over random equally important.
In Table~\ref{tab_cifar10_ipc1} to Table~\ref{tab_in1k_ipc50}, we report the LRS results under different $\lambda$.
A larger $\lambda$ gives higher priority to IOR, and a smaller $\lambda$ focuses more on HLR.
We encourage future newly proposed DD methods to enhance both HLR and IOR.
From Table~\ref{tab_cifar10_main_std} to Table~\ref{tab_in1k_main_std}, we provide the standard deviations for all benchmark results computed under 5 runs with different random seeds.

\begin{table}[h]
\centering
\tablestyle{8pt}{1.3}
\begin{tabular}{c|ccccc}
$\lambda$                         & 0.1  & 0.3  & 0.5  & 0.7  & 0.9  \\ \hline
\shline \rowcolor{hardlabel} DC   & 11.2 & 14.9 & 19.1 & 24.0 & 29.5 \\
\rowcolor{hardlabel} DSA          & 9.7 & 13.7 & 18.2 & 23.5 & 29.5 \\
\rowcolor{hardlabel} MTT          & 14.3 & 18.7 & 23.9 & 29.8 & 36.6 \\
\rowcolor{hardlabel} DM           & 9.0 & 12.8  & 17.0 & 22.0 & 27.6 \\
\rowcolor{hardlabel} DATADAM      & 11.9 & 15.8 & 20.2 & 25.2 & 30.9 \\
\shline \rowcolor{softlabel} DATM & 14.4 & 19.2 & 24.6 & 30.9 & 38.2 \\
\rowcolor{softlabel} SRe2L        & 7.0 & 10.4 & 14.3 & 18.8 & 23.9 \\
\rowcolor{softlabel} RDED         & 9.1 & 12.4  & 16.2 & 20.4 & 25.3  \\
\rowcolor{softlabel} D4M          & 11.4 & 14.7  & 18.4 & 22.6 & 27.3 
\end{tabular}
\caption{LRS evaluation results on CIFAR-10 IPC1 under different $\lambda$.}
\label{tab_cifar10_ipc1}
\end{table}

\begin{table}[h]
\centering
\tablestyle{8pt}{1.3}
\begin{tabular}{c|ccccc}
$\lambda$                         & 0.1  & 0.3  & 0.5  & 0.7  & 0.9  \\ \hline
\shline \rowcolor{hardlabel} DC   & 15.5 & 19.1 & 23.2 & 27.7 & 32.8 \\
\rowcolor{hardlabel} DSA          & 16.0 & 19.6 & 23.7 & 28.3 & 33.4 \\
\rowcolor{hardlabel} MTT          & 19.8 & 23.9 & 28.5 & 33.5 & 39.2 \\
\rowcolor{hardlabel} DM           & 14.7 & 18.2  & 22.2 & 26.7 & 31.6 \\
\rowcolor{hardlabel} DATADAM      & 16.1 & 19.7 & 23.8 & 28.4 & 33.5 \\
\shline \rowcolor{softlabel} DATM & 19.0 & 23.5 & 28.7 & 34.5 & 41.2 \\
\rowcolor{softlabel} SRe2L        & 7.3 & 10.4 & 13.8 & 17.7 & 22.1 \\
\rowcolor{softlabel} RDED         & 11.3 & 14.3  & 17.5 & 21.2 & 25.2  \\
\rowcolor{softlabel} D4M          & 14.3 & 17.4  & 20.8 & 24.6 & 28.7 
\end{tabular}
\caption{LRS evaluation results on CIFAR-10 IPC10 under different $\lambda$.}
\label{tab_cifar10_ipc10}
\end{table}

\begin{table}[h]
\centering
\tablestyle{8pt}{1.3}
\begin{tabular}{c|ccccc}
$\lambda$                         & 0.1  & 0.3  & 0.5  & 0.7  & 0.9  \\ \hline
\shline \rowcolor{hardlabel} DC   & 18.3 & 21.1 & 24.0 & 27.2 & 30.6 \\
\rowcolor{hardlabel} DSA          & 18.0 & 20.6 & 23.5 & 26.7 & 30.1 \\
\rowcolor{hardlabel} MTT          & 21.8 & 24.7 & 27.8 & 31.1 & 34.7 \\
\rowcolor{hardlabel} DM           & 18.7 & 21.4  & 24.3 & 27.5 & 30.9 \\
\rowcolor{hardlabel} DATADAM      & 19.8 & 22.6 & 25.6 & 28.8 & 32.3 \\
\shline \rowcolor{softlabel} DATM & 21.1 & 24.4 & 28.0 & 31.9 & 36.1 \\
\rowcolor{softlabel} SRe2L        & 8.3 & 11.2 & 14.4 & 18.0 & 22.0 \\
\rowcolor{softlabel} RDED         & 15.1 & 17.3  & 19.6 & 22.1 & 24.8  \\
\rowcolor{softlabel} D4M          & 17.9 & 20.3  & 22.8 & 25.4 & 28.3 
\end{tabular}
\caption{LRS evaluation results on CIFAR-10 IPC50 under different $\lambda$.}
\label{tab_cifar10_ipc50}
\end{table}

\begin{table}[t]
\centering
\tablestyle{8pt}{1.3}
\begin{tabular}{c|ccccc}
$\lambda$                         & 0.1  & 0.3  & 0.5  & 0.7  & 0.9   \\ \hline
\shline \rowcolor{hardlabel} DC   & 14.4 & 17.5 & 20.8 & 24.4 & 28.5  \\
\rowcolor{hardlabel} DSA          & 12.7 & 16.0 & 19.6 & 23.7 & 28.2  \\
\rowcolor{hardlabel} MTT          & 15.9 & 19.3 & 23.1 & 27.4 & 32.1  \\
\rowcolor{hardlabel} DM           & 12.1 & 15.3 & 18.9 & 22.8 & 27.2  \\
\rowcolor{hardlabel} DATADAM      & 12.9 & 16.2 & 19.9 & 23.9 & 28.5  \\
\shline \rowcolor{softlabel} DATM & 19.2 & 22.3 & 25.7 & 29.4 & 33.4  \\
\rowcolor{softlabel} SRe2L        & 10.8 & 13.6 & 16.7 & 20.2 & 24.0 \\
\rowcolor{softlabel} RDED         & 12.6 & 15.2 & 18.1 & 21.3 & 24.8  \\
\rowcolor{softlabel} D4M          & 16.9 & 19.7 & 22.7 & 25.9 & 29.5  
\end{tabular}
\caption{LRS evaluation results on CIFAR-100 IPC1 under different $\lambda$.}
\label{tab_cifar100_ipc1}
\end{table}

\begin{table}[t]
\centering
\tablestyle{8pt}{1.3}
\begin{tabular}{c|ccccc}
$\lambda$                         & 0.1  & 0.3  & 0.5  & 0.7  & 0.9   \\ \hline
\shline \rowcolor{hardlabel} DC   & 18.6 & 21.3 & 24.3 & 27.4 & 30.8  \\
\rowcolor{hardlabel} DSA          & 18.4 & 21.3 & 24.3 & 27.6 & 31.2  \\
\rowcolor{hardlabel} MTT          & 21.3 & 24.3 & 27.5 & 30.9 & 34.7  \\
\rowcolor{hardlabel} DM           & 17.1 & 19.9 & 23.0 & 26.2 & 29.8  \\
\rowcolor{hardlabel} DATADAM      & 18.6 & 21.5 & 24.6 & 28.0 & 31.7  \\
\shline \rowcolor{softlabel} DATM & 20.9 & 23.7 & 26.8 & 30.1 & 33.6  \\
\rowcolor{softlabel} SRe2L        & 11.0 & 12.9 & 15.0 & 17.3 & 19.8 \\
\rowcolor{softlabel} RDED         & 14.7 & 17.0 & 19.4 & 22.0 & 24.9  \\
\rowcolor{softlabel} D4M          & 14.3 & 17.4 & 20.9 & 24.7 & 29.0  
\end{tabular}
\caption{LRS evaluation results on CIFAR-100 IPC10 under different $\lambda$.}
\label{tab_cifar100_ipc10}
\end{table}

\begin{table}[h]
\centering
\tablestyle{8pt}{1.3}
\begin{tabular}{c|ccccc}
$\lambda$                         & 0.1  & 0.3  & 0.5  & 0.7  & 0.9   \\ \hline
\shline \rowcolor{hardlabel} DC   & 19.4 & 21.0 & 22.7 & 24.5 & 26.4  \\
\rowcolor{hardlabel} DSA          & 19.6 & 21.2 & 23.0 & 24.8 & 26.8  \\
\rowcolor{hardlabel} MTT          & 22.9 & 24.8 & 26.8 & 28.8 & 31.0  \\
\rowcolor{hardlabel} DM           & 21.3 & 23.1 & 24.9 & 26.9 & 29.0  \\
\rowcolor{hardlabel} DATADAM      & 22.9 & 24.8 & 26.8 & 28.9 & 31.1  \\
\shline \rowcolor{softlabel} DATM & 24.2 & 27.2 & 30.4 & 33.9 & 37.6  \\
\rowcolor{softlabel} SRe2L        & 12.1 & 14.1 & 16.2 & 18.5 & 21.0 \\
\rowcolor{softlabel} RDED         & 17.6 & 19.3 & 21.2 & 23.1 & 25.2  \\
\rowcolor{softlabel} D4M          & 18.3 & 21.1 & 24.2 & 27.5 & 31.1  
\end{tabular}
\caption{LRS evaluation results on CIFAR-100 IPC50 under different $\lambda$.}
\label{tab_cifar100_ipc50}
\end{table}

\begin{table}[t]
\centering
\tablestyle{8pt}{1.3}
\begin{tabular}{c|ccccc}
$\lambda$                         & 0.1  & 0.3  & 0.5  & 0.7  & 0.9  \\
\shline \rowcolor{hardlabel} DC   & 17.4 & 19.6 & 22.0 & 24.5 & 27.2  \\
\rowcolor{hardlabel} DSA          & 16.9 & 19.1 & 21.6 & 24.2 & 27.0  \\
\rowcolor{hardlabel} MTT          & 16.8 & 19.3 & 21.9 & 24.8 & 27.8 \\
\rowcolor{hardlabel} DM           & 15.0 & 17.5 & 20.2 & 23.1 & 26.2  \\
\shline \rowcolor{softlabel} DATM & 18.5 & 20.9 & 23.5 & 26.2 & 29.2 \\
\rowcolor{softlabel} EDF          & 18.4 & 20.9 & 23.5 & 26.3 & 29.4 \\
\rowcolor{softlabel} SRe2L        & 12.5 & 15.1 & 17.9 & 21.0 & 24.3 \\
\rowcolor{softlabel} RDED         & 15.8 & 18.3 & 20.9 & 23.8 & 26.9  \\
\rowcolor{softlabel} D4M          & 13.8 & 16.1 & 18.6 & 21.2 & 24.1
\end{tabular}
\caption{LRS evaluation results on TinyImageNet IPC1 under different $\lambda$.}
\label{tab_tiny_ipc1}
\end{table}

\begin{table}[t]
\centering
\tablestyle{8pt}{1.3}
\begin{tabular}{c|ccccc}
$\lambda$                         & 0.1  & 0.3  & 0.5  & 0.7  & 0.9  \\
\shline \rowcolor{hardlabel} DC   & 19.7 & 21.7 & 23.9 & 26.3 & 28.7  \\
\rowcolor{hardlabel} DSA          & 20.0 & 22.0 & 24.1 & 26.3 & 28.7  \\
\rowcolor{hardlabel} MTT          & 21.9 & 24.2 & 26.7 & 29.3 & 32.1 \\
\rowcolor{hardlabel} DM           & 18.2 & 20.6 & 23.1 & 25.8 & 28.7  \\
\shline \rowcolor{softlabel} DATM & 20.9 & 23.4 & 26.0 & 28.8 & 31.8 \\
\rowcolor{softlabel} EDF          & 20.9 & 23.5 & 26.2 & 29.2 & 32.3 \\
\rowcolor{softlabel} SRe2L        & 12.8 & 14.9 & 17.1 & 19.5 & 22.1 \\
\rowcolor{softlabel} RDED         & 18.2 & 20.1 & 22.1 & 24.2 & 26.5  \\
\rowcolor{softlabel} D4M          & 15.1 & 16.9 & 18.9 & 21.1 & 23.3
\end{tabular}
\caption{LRS evaluation results on TinyImageNet IPC10 under different $\lambda$.}
\label{tab_tiny_ipc10}
\end{table}

\begin{table}[t]
\centering
\tablestyle{8pt}{1.3}
\begin{tabular}{c|ccccc}
$\lambda$                         & 0.1  & 0.3  & 0.5  & 0.7  & 0.9  \\
\shline \rowcolor{hardlabel} DC   & 19.4 & 20.8 & 22.2 & 23.7 & 25.2  \\
\rowcolor{hardlabel} DSA          & 20.9 & 22.8 & 24.8 & 26.9 & 29.1  \\
\rowcolor{hardlabel} MTT          & 21.7 & 23.7 & 25.8 & 28.0 & 30.3 \\
\rowcolor{hardlabel} DM           & 20.4 & 22.2 & 24.1 & 26.1 & 28.1  \\
\shline \rowcolor{softlabel} DATM & 22.6 & 24.9 & 27.2 & 29.8 & 32.4 \\
\rowcolor{softlabel} EDF          & 22.5 & 24.9 & 27.3 & 30.0 & 32.8 \\
\rowcolor{softlabel} SRe2L        & 15.5 & 17.0 & 18.6 & 20.3 & 22.1 \\
\rowcolor{softlabel} RDED         & 21.4 & 22.5 & 23.7 & 24.8 & 26.0  \\
\rowcolor{softlabel} D4M          & 17.9 & 20.8 & 23.8 & 27.2 & 30.8
\end{tabular}
\caption{LRS evaluation results on TinyImageNet IPC50 under different $\lambda$.}
\label{tab_tiny_ipc50}
\end{table}

\begin{table}[t]
\centering
\tablestyle{8pt}{1.3}
\begin{tabular}{c|ccccc}
$\lambda$                          & 0.1  & 0.3  & 0.5  & 0.7  & 0.9   \\
\shline \rowcolor{softlabel} SRe2L & 9.9 & 12.9 & 16.2 & 19.9 & 24.0 \\
\rowcolor{softlabel} RDED          & 10.2 & 13.3 & 16.8 & 20.8 & 25.2  \\
\rowcolor{softlabel} D4M           & 10.1 & 13.1 & 16.4 & 20.2 & 24.4  \\
\rowcolor{softlabel} DWA           & 10.0 & 13.0 & 16.3 & 20.0 & 24.1  \\
\rowcolor{softlabel} CDA           & 9.9 & 12.8 & 16.1 & 19.7 & 23.7  \\
\rowcolor{softlabel} EDC           & 10.1 & 13.1 & 16.4 & 20.1 & 24.3  \\
\rowcolor{softlabel} G-VBSM        & 10.0 & 12.9 & 16.3 & 20.0 & 24.1 
\end{tabular}
\caption{LRS evaluation results on ImageNet1K IPC1 under different $\lambda$.}
\label{tab_in1k_ipc1}
\end{table}

\begin{table}[t]
\centering
\tablestyle{8pt}{1.3}
\begin{tabular}{c|ccccc}
$\lambda$                          & 0.1  & 0.3  & 0.5  & 0.7  & 0.9   \\
\shline \rowcolor{softlabel} SRe2L & 9.9 & 12.0 & 14.2 & 16.7 & 19.3 \\
\rowcolor{softlabel} RDED          & 11.4 & 14.2 & 17.4 & 20.8 & 24.6  \\
\rowcolor{softlabel} D4M           & 10.6 & 13.0 & 15.8 & 18.7 & 22.0  \\
\rowcolor{softlabel} DWA           & 10.3 & 13.1 & 16.1 & 19.5 & 23.2  \\
\rowcolor{softlabel} CDA           & 10.1 & 12.6 & 15.3 & 18.3 & 21.6  \\
\rowcolor{softlabel} EDC           & 11.0 & 13.9 & 17.1 & 20.6 & 24.6  \\
\rowcolor{softlabel} G-VBSM        & 10.1 & 12.7 & 15.5 & 18.6 & 22.1 
\end{tabular}
\caption{LRS evaluation results on ImageNet1K IPC10 under different $\lambda$.}
\label{tab_in1k_ipc10}
\end{table}

\begin{table}[t]
\centering
\tablestyle{8pt}{1.3}
\begin{tabular}{c|ccccc}
$\lambda$                          & 0.1  & 0.3  & 0.5  & 0.7  & 0.9   \\
\shline \rowcolor{softlabel} SRe2L & 10.3 & 12.5 & 14.8 & 17.4 & 20.2 \\
\rowcolor{softlabel} RDED          & 14.0 & 16.2 & 18.6 & 21.2 & 23.9  \\
\rowcolor{softlabel} D4M           & 12.9 & 15.1 & 17.6 & 20.2 & 23.0  \\
\rowcolor{softlabel} DWA           & 11.3 & 13.7 & 16.3 & 19.1 & 22.1  \\
\rowcolor{softlabel} CDA           & 10.8 & 13.3 & 16.1 & 19.1 & 22.4  \\
\rowcolor{softlabel} EDC           & 13.7 & 16.2 & 18.9 & 21.9 & 25.1  \\
\rowcolor{softlabel} G-VBSM        & 12.6 & 14.9 & 17.4 & 20.0 & 22.9 
\end{tabular}
\caption{LRS evaluation results on ImageNet1K IPC50 under different $\lambda$.}
\label{tab_in1k_ipc50}
\end{table}

\begin{table}[t]
    \centering
    \begin{subtable}{1.0\textwidth}
        \tablestyle{8pt}{1.3}
        \begin{tabular}{c|cccccccccc}
        ipc        & \multicolumn{3}{c}{1} & \multicolumn{3}{c}{10} & \multicolumn{3}{c}{50} \\
        metric        & HLR$\downarrow$   & IOR$\uparrow$   & LRS$\uparrow$   & HLR$\downarrow$   & IOR$\uparrow$   & LRS$\uparrow$   & HLR$\downarrow$   & IOR$\uparrow$   & LRS$\uparrow$   \\ \shline \rowcolor{hardlabel}
        DC         &  0.2     &  0.2     &  0.3     & 0.3    &  0.2     & 0.3      &  0.2      &  0.4     &  0.3     \\ \rowcolor{hardlabel}
        DSA        &  0.3     &  0.3     & 0.2      & 0.4   & 0.2      & 0.2      &  0.5    & 0.4      & 0.4      \\ \rowcolor{hardlabel}
        MTT        &  0.5     &  0.7     & 0.6      & 0.6   & 0.8      & 0.8      &  0.5    &  0.4     & 0.5      \\ \rowcolor{hardlabel}
        DM         &  0.7     &  0.6     &  0.5     &  0.8   & 0.9      & 1.0      & 0.7       &  0.7     & 0.7      \\ \rowcolor{hardlabel}
        DATADAM    &  0.8     &  0.5     & 0.6      &  0.6  & 0.6      & 0.5      & 0.7     & 0.5      &  0.6     \\ \shline \rowcolor{softlabel}
        DATM       &  0.7     &  0.4     & 0.8      & 0.5   & 0.7   &  0.6     &  0.3      & 0.7      &  0.5     \\ \rowcolor{softlabel}
        SRe2L      &  0.5     &  0.6    & 0.6      & 0.8   & 0.8   &  0.6     &  0.5    & 0.8    & 0.7      \\ \rowcolor{softlabel}
        RDED       &  0.7     &  0.9     & 0.7      &  0.8   & 0.7    &  0.8     &  0.9      & 1.2      & 0.9      \\ \rowcolor{softlabel}
        D4M        &  0.8     &  0.8     &  0.6     & 0.7    & 0.9    &  0.9     & 0.8    &  1.0     &  0.9    
        \end{tabular}
        \caption{Standard deviations of LRS results on CIFAR-10.}
        \label{tab_cifar10_main_std}
    \end{subtable}

    \begin{subtable}{1.0\textwidth}
        \centering
        \tablestyle{8pt}{1.3}
        \begin{tabular}{c|cccccccccc}
        ipc        & \multicolumn{3}{c}{1} & \multicolumn{3}{c}{10} & \multicolumn{3}{c}{50} \\
        metric        & HLR$\downarrow$   & IOR$\uparrow$   & LRS$\uparrow$   & HLR$\downarrow$   & IOR$\uparrow$   & LRS$\uparrow$   & HLR$\downarrow$   & IOR$\uparrow$   & LRS$\uparrow$   \\ \shline \rowcolor{hardlabel}
        DC         &  0.4     &  0.3     &  0.3     & 0.3    &  0.5     & 0.4      &  0.2      &  0.6     &  0.7     \\ \rowcolor{hardlabel}
        DSA        &  0.5     &  0.5     & 0.5      & 0.4   & 0.4      & 0.3      &  0.4    & 0.5      & 0.4      \\ \rowcolor{hardlabel}
        MTT        &  0.5     &  0.6     & 0.5      & 0.7   & 0.7      & 0.6      &  0.5    &  0.6     & 0.6      \\ \rowcolor{hardlabel}
        DM         &  0.6     &  0.8     &  0.7     &  0.9   & 0.9      & 0.9      & 0.7   &  0.7     & 0.5      \\ \rowcolor{hardlabel}
        DATADAM    &  0.6     &  0.7     & 0.8      &  0.5  & 0.8      & 0.7      & 0.7     & 0.6      &  0.6     \\ \shline \rowcolor{softlabel}
        DATM       &  0.7     &  0.5     & 0.6      & 0.6   & 0.6   &  0.6     &  0.5      & 0.8      &  0.7     \\ \rowcolor{softlabel}
        SRe2L      &  1.1     &  0.9    & 0.9      & 0.8   & 0.7   &  0.7     &  0.5    & 0.9    & 0.7      \\ \rowcolor{softlabel}
        RDED       &  0.7     &  1.0     & 0.8      &  0.6   & 0.9    &  0.6     &  0.8      & 1.1      & 0.9      \\ \rowcolor{softlabel}
        D4M        &  0.5     &  0.6     &  0.4     & 0.7    & 0.8    &  0.6     & 0.8    &  0.9     &  0.9    
        \end{tabular}
        \caption{Standard deviations of LRS results on CIFAR-100.}
        \label{tab_cifar100_main_std}
    \end{subtable}

    \begin{subtable}{1.0\textwidth}
        \centering
        \tablestyle{8pt}{1.3}
        \begin{tabular}{c|cccccccccc}
        ipc        & \multicolumn{3}{c}{1} & \multicolumn{3}{c}{10} & \multicolumn{3}{c}{50} \\
        metric        & HLR$\downarrow$   & IOR$\uparrow$   & LRS$\uparrow$   & HLR$\downarrow$   & IOR$\uparrow$   & LRS$\uparrow$   & HLR$\downarrow$   & IOR$\uparrow$   & LRS$\uparrow$   \\ \shline \rowcolor{hardlabel}
        DC         &  0.4     &  0.3     &  0.3     & 0.2    &  0.5    & 0.5      &  0.4      &  0.6     &  0.4     \\ \rowcolor{hardlabel}
        DSA        &  0.5     &  0.4     & 0.6      & 0.3   & 0.7      & 0.4      &  0.5    & 0.5      & 0.4      \\ \rowcolor{hardlabel}
        MTT        &  0.5     &  0.6     & 0.6      & 0.4   & 0.7      & 0.6      &  0.3    &  0.8     & 0.6      \\ \rowcolor{hardlabel}
        DM         &  0.3     &  0.7     &  0.5     &  0.8   & 0.9      & 0.8      & 0.5       &  0.8     & 0.7      \\ \rowcolor{hardlabel} \shline \rowcolor{softlabel}
        DATM       &  0.5     &  0.4     & 0.4      & 0.3   & 0.5   &  0.6     &  0.4      & 0.9      &  0.6     \\ \rowcolor{softlabel}
        SRe2L      &  0.7     &  0.7    & 0.7      & 0.5   & 0.4   &  0.5    &  0.4    & 0.8    & 0.8      \\ \rowcolor{softlabel}
        RDED       &  0.6     &  0.8     & 0.7      &  0.5   & 0.9    &  0.7     &  0.7      & 1.0      & 0.9      \\ \rowcolor{softlabel}
        D4M        &  0.6     &  0.7     &  0.6     & 0.7    & 0.8    &  0.6     & 0.9    &  1.1     &  0.8    
        \end{tabular}
        \caption{Standard deviations of LRS results on TinyImageNet.}
        \label{tab_cifar10_main_std}
    \end{subtable}

    \begin{subtable}{1.0\textwidth}
        \centering
        \tablestyle{8pt}{1.3}
        \begin{tabular}{c|ccccccccc}
        ipc           & \multicolumn{3}{c}{1} & \multicolumn{3}{c}{10} & \multicolumn{3}{c}{50} \\
        metric        & HLR$\downarrow$   & IOR$\uparrow$   & LRS$\uparrow$   & HLR$\downarrow$   & IOR$\uparrow$   & LRS$\uparrow$   & HLR$\downarrow$   & IOR$\uparrow$   & LRS$\uparrow$   \\ \shline \rowcolor{softlabel}
        SRe2L    &  0.6     &  1.1     &  0.8     &  0.5      &  0.8     &  0.9     &  0.7      &  1.0     & 0.7  \\ \rowcolor{softlabel}
        RDED     &  0.5     &  0.7     &  0.6     &  0.4      &  0.9     &  0.8     &  0.7      &  0.7     & 0.7  \\ \rowcolor{softlabel}
        D4M      &  0.8     &  0.6     &  0.6     &  0.5      & 1.1  & 0.9     &  0.6      &  0.8     &  0.5 \\ \rowcolor{softlabel}
        DWA      &  0.8     &  1.2     &  1.1     &  0.9      &  1.1     &    1.0   &  0.7      &  0.8     &  0.9 \\ \rowcolor{softlabel}
        CDA      &  0.6     &  0.6     &  0.8     &  0.7      &  0.8     &   0.7  &   0.4     &  0.8    &  0.6 \\ \rowcolor{softlabel}
        EDC      &   0.3    &   0.8    &  0.4    &  0.5    &   0.4   &  0.5   &   0.5    &   1.1    & 0.9  \\ \rowcolor{softlabel}
        G-VBSM   &   0.6    &   1.2   &  1.0     &   0.5    &   1.3   &  0.9  &  0.6    &  0.9    &  0.8 \\ 
        \end{tabular}
        \caption{Standard deviations of LRS results on ImageNet-1K.}
        \label{tab_in1k_main_std}
    \end{subtable}
\end{table}

\section{Additional Related Work}

We acknowledge that DD-Ranking has not included enough dataset distillation methods. We discuss them here. In the near future, we will continue to extend our benchmark and include more baseline methods. 
\begin{table}[t]
\centering
\tablestyle{10pt}{1.2}
\begin{tabular}{c|c}
Category  & Method   \\ \shline
\multicolumn{1}{l|}{\multirow{5}{*}{Kernel-based}} & KIP-FC~\citep{KIP-FC}       \\
                                                & KIP-ConvNet~\citep{KIP-ConvNet}      \\
                                                & FRePo~\citep{FRePo}      \\
                                                & RFAD~\citep{RFAD}    \\ 
                                                & RCIG~\citep{RCIG}   \\ \shline

\multicolumn{1}{l|}{\multirow{4}{*}{Gradient-matching}}                          & DC~\citep{zhao2021dataset}       \\
                                                            & DSA~\citep{zhao2021dsa}      \\
                                                            & DCC~\citep{DCC}      \\
                                                            & LCMat~\citep{LCMat}    \\ \shline
\multicolumn{1}{l|}{\multirow{10}{*}{Trajectory-matching}}                        & MTT~\citep{cazenavette2022dataset}      \\
                                                            & TESLA~\citep{cui2023scaling}    \\
                                                            & FTD~\citep{du2023minimizing}     \\
                                                            & SeqMatch~\citep{Du2023SequentialSM} \\
                                                            & DATM~\citep{guo2022deepcore}     \\
                                                            & ATT~\citep{ATT}      \\
                                                            & NSD~\citep{NSD}      \\
                                                            & PAD~\citep{li2024prioritizealignmentdatasetdistillation}      \\
                                                            
                                                            & EDF~\citep{wang2025emphasizingdiscriminativefeaturesdataset}      \\
                                                            & SelMatch~\citep{lee2024selmatcheffectivelyscalingdataset} \\ \shline
\multicolumn{1}{l|}{\multirow{6}{*}{Distribution-matching}} & DM~\citep{zhao2021datasetdm}       \\
\multicolumn{1}{l|}{}                                       & CAFE~\citep{Wang2022CAFELT}     \\

\multicolumn{1}{l|}{}                                       & IDM~\citep{zhao2023improved}      \\
\multicolumn{1}{l|}{}                                       & DREAM~\citep{Liu2023DREAMED}    \\
\multicolumn{1}{l|}{}                                       & M3D~\citep{M3D}      \\ 
\multicolumn{1}{l|}{}                                       & NCFD~\citep{wang2025datasetdistillationneuralcharacteristic}      \\ \shline
\multicolumn{1}{l|}{\multirow{7}{*}{Generative model}}      & DiM~\cite{Wang2023DiMDD}       \\
\multicolumn{1}{l|}{}                                       & GLaD~\citep{cazenavette2023generalizing}     \\
\multicolumn{1}{l|}{}                                       & H-PD~\citep{H-GLaD}      \\
\multicolumn{1}{l|}{}                                       & LD3M~\citep{Moser2024LatentDD}    \\
\multicolumn{1}{l|}{}                                       & IT-GAN~\citep{zhao2022synthesizinginformativetrainingsamples}    \\
\multicolumn{1}{l|}{}                                       & D4M~\cite{D4M}    \\
\multicolumn{1}{l|}{}                                       & Minimax Diffusion~\cite{Gu2023EfficientDD}    \\ \shline
\multicolumn{1}{l|}{\multirow{7}{*}{Decoupled}} & SRe2L~\citep{yin2024squeeze}      \\
\multicolumn{1}{l|}{}                                        & RDED~\citep{sun2024diversity}     \\
\multicolumn{1}{l|}{}                                        & HeLIO~\citep{yu2024heavylabelsoutdataset}         \\ 
\multicolumn{1}{l|}{}     &  DWA~\citep{du2024diversitydrivensynthesisenhancingdataset}  \\ 
\multicolumn{1}{l|}{}     &     CDA~\citep{yin2023datasetdistillationlargedata} \\ 
\multicolumn{1}{l|}{}     &     EDC~\citep{shao2024elucidating} \\ 
\multicolumn{1}{l|}{}     &     G-VBSM~\citep{shao2024generalized} \\ \shline
\multicolumn{1}{l|}{\multirow{3}{*}{Others}} & MIM4DD~\citep{shang2023mim4ddmutualinformationmaximization}       \\
\multicolumn{1}{l|}{}                                        & DQAS~\citep{zhao2024datasetquantizationactivelearning}     \\
\multicolumn{1}{l|}{}                                        & LDD~\citep{zhao2025distillinglongtaileddatasets}         \\
\end{tabular}
\caption{Summary of previous works on dataset distillation}
\label{more_rel_work}
\end{table}



\end{document}